\documentclass[letterpaper]{article} 
\usepackage[preprint]{aaai2027}  
\usepackage[hyphens]{url}  
\usepackage{graphicx} 
\usepackage{natbib}  
\usepackage{caption} 
\usepackage{algorithm}
\usepackage{algorithmic}
\usepackage{booktabs}
\usepackage{amsmath}
\usepackage{amssymb}
\usepackage{multirow}
\usepackage{placeins}

\title{Operationally Guided Placement-Aware Learning for Industrial Online 3D Bin Packing}
\author {
    Dheeraj Poolavaram\textsuperscript{\rm 1}\corresponding,
    Aanchal Rajesh Chugh\textsuperscript{\rm 1},
    Sebastian Dorn\textsuperscript{\rm 1}
}
\affiliations {
    \textsuperscript{\rm 1}TTZ Landsberg am Lech, Technische Hochschule Augsburg, Augsburg, Germany\\

    {\{dheeraj.poolavaram, aanchal.rajesh.chugh, sebastian.dorn\}@tha.de}}

\begin{document}

\maketitle

\begin{abstract}

The online three-dimensional bin packing problem (3D-BPP) is a longstanding challenge in logistics and industrial palletizing. Recent learning-based methods use a learned policy to select among feasible candidate placements. Performance depends on the candidate generator and representation, especially in industrial settings where packings must be space-efficient, stable, compact, and balanced. However, prior work has mainly optimized the policy, while candidate generation and representation remain largely geometry-driven. We address this gap with OPAL, an operationally guided placement-aware learning framework for industrial online 3D-BPP which combines an Operationally Guided Empty-Maximal-Space generator (OG-EMS), an operational representation for each candidate placement, and a masked ranking policy trained with proximal policy optimization. OG-EMS evaluates multiple anchors within each free-space region and prioritizes low, well-supported, compact, and spatially diverse placements. An xLSTM-based Placement Encoder models dependencies among geometric and operational candidate attributes, while a lightweight recurrent core combines the resulting embeddings with the current item and pallet state to rank feasible actions. On the BED-BPP benchmark, OPAL achieves a mean space utilization of 0.49, with improvements of 15.1\% from operationally guided candidate generation and 6.3\% from learned ranking, while maintaining robust inference-time performance.

\end{abstract}

\section{Introduction}

The three-dimensional bin packing problem (3D-BPP) is a fundamental combinatorial optimization challenge with direct relevance to palletization, warehouse logistics, and container loading \citep{martello2000three,bortfeldt2013constraints}. In the online variant, items are processed sequentially and each placement is committed without subsequent rearrangement \citep{ha2017online,zhao2021online,zhao2022pct}. In industrial palletizing, volume utilization alone is insufficient. A deployable packing must also maintain support, avoid excessive overhang, limit stack-load risk, preserve a reasonable center of gravity, and remain reachable for handling \citep{bortfeldt2013constraints}. This work targets industrial order packing on standardized Euro pallets \citep{epal_euro_pallet}.

A common design in learning-based online packing avoids direct continuous placement prediction. Instead, a geometric generator proposes a finite set of feasible or near-feasible placements, a mask removes invalid actions, and a learned policy ranks the remaining candidates \citep{zhao2021online,zhao2022pct,xiong2024gopt}. We refer to this combination of candidate generation, candidate representation, feasibility masking, and learned ranking as the candidate-selection interface. This design keeps the neural model within a structured planning loop. However, it also makes overall performance strongly dependent on how this interface is constructed, an aspect that prior work has under-emphasized. Thus, performance depends on both exposure, which placements are generated, and ranking, whether their representation lets the policy distinguish them.

We combine operationally guided candidate generation with placement-aware candidate representation and ranking in OPAL, an Operationally Guided Placement-Aware Learning framework built around a masked candidate-selection formulation. The candidate generator, Operationally Guided EMS (OG-EMS), starts from Empty-Maximal-Space (EMS) regions \citep{parreno2008maximal}, evaluates multiple anchors per region, and scores them with low-height, wall-contact, fit-quality, sliver-avoidance, corner, support, and diversity priors. Each anchor-orientation action is encoded as its own 15-dimensional vector containing EMS position and dimensions, the item footprint under that orientation, bottom support, support-region centering margin, normalized top height, stack-load ratio, fragility feasibility, side support, and placement-effort score. The ranker embeds each candidate with a Placement Encoder (PE) built on an xLSTM stack \citep{beck2024xlstm}, mixes item and candidate tokens with a recurrent action mixer (LRAM) \citep{schmied2025largerecurrentactionmodel}, and scores feasible candidates through an actor head. OPAL therefore improves both which placements are exposed to the policy and how effectively the policy can distinguish among them.

Beyond the core comparisons, we evaluate a shallow inference-time lookahead and analyze how its gains vary with baseline packing efficiency. Experiments across pallet footprints test whether the relative ordering of Base-EMS and OG-EMS can be generalized beyond the primary Euro-pallet setting. We further assess operational quality through density, surface and side support, center-of-gravity balance, and end-to-end inference latency.

\noindent The main contributions are as follows:
\begin{itemize}

\item OG-EMS, an operationally guided multi-anchor candidate generator, paired with a 15-dimensional candidate representation and placement-aware learned ranking.

\item An interface-first analysis of masked candidate-selection packing on 1500 real-world orders from the BED-BPP dataset \citep{kagerer2023bedbpp}, examining the roles of candidate generation, representation, ranking architecture, and temporal recurrence.

\item An evaluation beyond volume utilization, including operational KPI trade-offs, inference-time lookahead, and pallet-footprint sensitivity.
\end{itemize}

\section{Related Work}

\paragraph{Classical and industrial 3D bin packing.}
Research on the 3D-BPP and container loading spans exact methods, constructive heuristics, maximal-space methods, extreme-point rules, and metaheuristics. Classical formulations establish the combinatorial difficulty of packing cuboidal items under geometric feasibility constraints \citep{martello2000three}. Industrial container loading additionally requires stability, support, load-bearing, and handling constraints that utilization alone does not capture \citep{bortfeldt2013constraints}. Many online systems reduce continuous placement to a finite set of geometrically meaningful candidates. Extreme-point and maximal-space methods construct these candidates while preserving the spatial structure of the remaining free space
\citep{crainic2008extreme,parreno2008maximal,ha2017online}. OPAL retains explicit geometric candidate generation and hard feasibility checks, and uses learning only to rank a bounded feasible set.

\paragraph{Learning-based online 3D bin packing.}
Early deep-learning approaches framed packing as sequence prediction \citep{hu2017solving} or attention-based construction \citep{zhang2021attend2pack}. Constrained deep reinforcement learning with feasibility-masked action spaces established the masked-selection paradigm for online 3D-BPP \citep{zhao2021online}. PCT introduced packing-configuration trees as a structured state-action representation \citep{zhao2022pct}, while GOPT used a Transformer policy to rank finite placement subspaces generated from the packing state \citep{xiong2024gopt}.

More recent methods explicitly structure the decision or planning process. ASAP decomposes decision making into learned pruning and selection policies to improve generalization \citep{fang2025asap}, DeliPacker extends packing-configuration representations toward deliberate planning over structured trees \citep{zhao2026delipacker}, and MPC-3D-BP uses short-horizon lookahead parcels with Monte Carlo tree search for model-predictive planning under partially visible future items \citep{fang2026lookahead}. ASAP and MPC-3D-BP extend PCT- and GOPT-based pipelines at the decision or planning stage while leaving candidate generation and representation unchanged. They could therefore, in principle, be layered on top of the candidate interface proposed in this work.

\paragraph{KPI-guided hybrid methods.}
Hybrid and multi-objective methods have similarly incorporated practical packing criteria beyond volumetric utilization. Hybrid genetic approaches combine constructive placement with evolutionary refinement while accounting for stability and fragility constraints \citep{ancora2020hybrid}, while multi-objective formulations consider load balance, stability, and product-family requirements alongside conventional packing objectives \citep{erbayrak2021multiobjective}. GENPACK follows this direction with a KPI-guided genetic-algorithm pipeline that combines constructive initialization, genetic refinement, and post-processing while explicitly optimizing industrial indicators including density, support, and balance \citep{poolavaram2026genpack}. It represents a state-of-the-art approach to industrial bin packing and is included as a contextual reference because it explicitly optimizes industrial KPIs such as density, support, and balance.

\section{Problem Formulation}
\label{sec:problemformulation}

An order contains $T$ rectangular items with dimensions, weight, and
handling attributes. Before packing, the items are sorted by descending
footprint area, yielding the fixed sequence
\begin{equation}
\mathcal{O}=(i_1,i_2,\ldots,i_T).
\end{equation}
The presorting procedure is analogous to the CUT-1 and CUT-2 sequences used by PCT and GOPT \citep{zhao2022pct,xiong2024gopt}. After sequencing, the policy observes the current pallet state and item, commits each placement irrevocably, and performs no rearrangement or post-processing in the primary configuration. We therefore study online placement decisions under a precomputed item sequence rather than unknown-order arrivals.

At step $t$, the system observes the pallet state $P_t$ and the current item $i_t$. Let $K$ denote the per-step budget on retained
region-anchor records, each contributing up to two orientation-specific candidate rows, and let $N_t\leq 2K$ denote the number of candidate placements generated at step $t$. The resulting candidate set is
\begin{equation}
\mathcal{C}_t=\{a_{t,1},\ldots,a_{t,N_t}\},
\end{equation}
with a corresponding feasibility mask $M_t\in\{0,1\}^{2K}$. Each candidate action $a_{t,j}$ specifies a placement position and an orientation for item $i_t$. The entry $M_{t,j}=1$ marks row $j$ as admissible, meaning the placement is geometrically valid and satisfies the operational constraints of Section~\ref{sec:ogems}; $M_{t,j}=0$ marks it inadmissible, covering constraint-violating placements and the $2K-N_t$ padding rows, which are excluded from sampling and selection. The policy commits one admissible action
\begin{equation}
a_t \in \{a_{t,j}\in \mathcal{C}_t:M_{t,j}=1\},
\end{equation}
where $a_t$ is the placement executed at step $t$: the position and orientation at which $i_t$ is irrevocably placed. Afterwards, the environment updates the heightmap, the placed-item set, the support structure, and the remaining order. A masked action can be neither selected nor recovered later in the order.

The stochastic policy $\pi_\theta$ is the actor--critic network of Section~\ref{sec:method} with trainable parameters $\theta$, learned by maximizing the expected discounted cumulative shaped reward over an order:
\begin{equation}
\theta^\star
=
\arg\max_\theta
\mathbb{E}_{\pi_\theta}
\left[
\sum_{t=1}^{T}\gamma^{t-1}r_t
\right],
\end{equation}
where $\theta^\star$ denotes the resulting optimized parameter vector,
$\gamma\in[0,1]$ is the discount factor, and the expectation is taken
over trajectories induced by $\pi_\theta$ and the environment.

For a successful placement, the step reward is

\begin{equation}
\begin{split}
r_t ={}& w_f\Delta f_t + w_s S_t + w_b B_t
- w_\tau \tau_t - w_r R_t
- w_h \chi_t^{\mathrm{height}} \\
&{} - w_g G_t
+ w_u U_t
+ w_w \chi_t^{\mathrm{wall}}
+ w_{ss}\chi_t^{\mathrm{side}},
\end{split}
\end{equation}
where $\Delta f_t$ is the increase in raw absolute density, $S_t$ is the bottom-support fraction, $B_t$ is the support-region centering term, $\tau_t$ is the placement-effort score, and $R_t$ combines unsupported footprint and excess stack-load risk. The terms $\chi_t^{\mathrm{height}}$ and $G_t$ penalize fill-weighted placement height and fill-weighted growth of the pallet maximum height, respectively; $U_t$ rewards low placements, $\chi_t^{\mathrm{wall}}$ rewards proximity to a pallet boundary, and $\chi_t^{\mathrm{side}}$ measures lateral support from previously placed items. Exact definitions are provided in Supplementary
Material~\ref{app:reward_definitions}. These local training signals are distinct from the final-layout operational KPIs defined in
Supplementary Material~\ref{app:kpi_definitions}.

Henceforth, we use the reward weights
\begin{equation}
\begin{split}
&(w_f,w_s,w_b,w_\tau,w_r,w_h,w_g,w_u,w_w,w_{ss})\\
&\qquad=(1.25,\allowbreak\,0.45,\allowbreak\,0.30,\allowbreak\,0.08,
\allowbreak\,0.35,\allowbreak\,0.90,\allowbreak\,1.80,\\
&\qquad\phantom{=(}0.60,\allowbreak\,0.20,\allowbreak\,0.15),
\end{split}
\end{equation}
which were selected through preliminary ablations and fixed
before evaluation on the reporting orders. 

\paragraph{Candidate exposure and ranking.}
Because the policy can select only from $\mathcal{C}_t$, a high-quality placement omitted during generation cannot be recovered by the ranker, which motivates treating candidate generation and representation as separate parts of the policy interface. A controlled generator comparison using a common deterministic selector is reported in Supplementary Material~\ref{app:generator_exposure}.

\section{Method}
\label{sec:method}

\subsection{Overview}

\begin{figure*}[t]
\centering
\includegraphics[width=\textwidth]{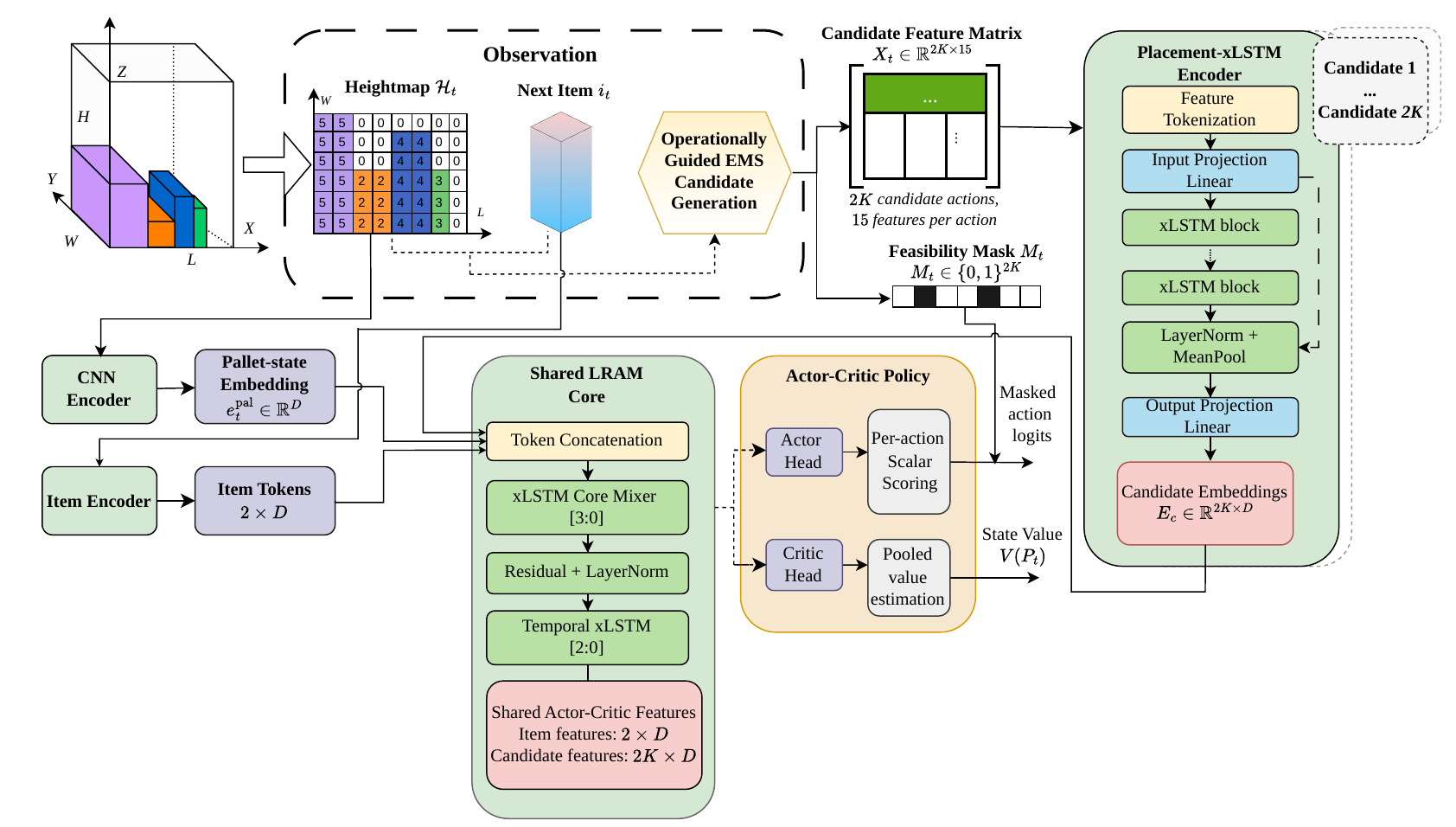}
\caption{Overview of OPAL. From the current pallet heightmap and next item, OG-EMS retains up to $K$ region-anchor records, where $K$ is the configured record budget, and represents up to two admissible orientations per record as at most $2K$ candidate actions. Each action is described by a 15-dimensional feature vector, giving $X_t\in\mathbb{R}^{2K\times15}$, together with the aligned feasibility mask $M_t\in\{0,1\}^{2K}$. The Placement-xLSTM Encoder independently maps each action row to a $D$-dimensional embedding, producing $E_c\in\mathbb{R}^{2K\times D}$, while a compact convolutional encoder maps the pallet state $P_t$ to the pallet-state embedding $e_t^{\mathrm{pal}}\in\mathbb{R}^{D}$. Candidate embeddings, orientation-conditioned item tokens, and the pallet-state embedding are mixed by the shared LRAM core and, when enabled, the temporal xLSTM. For each action, the actor combines the mixed candidate representation with its orientation-matched item token and produces one action logit; infeasible and padded rows are masked before sampling. The critic pools the shared representations to estimate the state value $V(P_t)$.}

\label{fig:architecture}

\end{figure*}

OPAL addresses both components of the candidate interface: candidate exposure and candidate representation. OG-EMS determines which actions are exposed, while the candidate representation and Placement Encoder determine what information is available to the policy when ranking them. Figure~\ref{fig:architecture} shows the resulting decision pipeline. The complete online packing procedure is summarized in Supplementary Material~\ref{app:algorithm}.

\subsection{Operationally Guided EMS (OG-EMS)}
\label{sec:ogems}

Base-EMS enumerates empty maximal spaces and prioritizes the resulting placement candidates using simple geometric criteria. This is efficient, but it may expose many placements that are feasible yet
operationally undesirable. OG-EMS starts from the same EMS regions but evaluates multiple anchor patterns and admissible orientations within each region. It then constructs a bounded candidate set using a geometric exposure cost, support-based ordering, and spatial-diversity filtering, stated in full in Supplementary Material~\ref{app:ogems_selection}. The learned policy ranks only the actions retained in this exposed set.

\paragraph{Generation.}
For each EMS region, OG-EMS evaluates five anchor patterns: the four bottom corners and the region center. At each pattern, both allowed item orientations are instantiated and checked for geometric validity and physical support stability. Each eligible placement is evaluated using the five-term geometric exposure cost
\begin{equation}
C_{\mathrm{OG}}
=
\beta_h z
+
\beta_w d_w
+
\beta_{\mathrm{res}}{\mathrm{res}}
+
\beta_v v
+
\beta_c d_c,
\end{equation}
where $z$, $d_w$, $\mathrm{res}$, $v$, and $d_c$ denote lower placement
height, nearest-wall distance, horizontal residual EMS slack,
narrow-slack penalty, and nearest-corner distance, respectively. Their
exact definitions and grid scaling are provided in Supplementary
Material~\ref{app:ogems_cost}. We found
$(\beta_h,\beta_w,\beta_{\mathrm{res}},\beta_v,\beta_c)
=
(5.0,\allowbreak\,2.0,\allowbreak\,0.8,\allowbreak\,3.0,
\allowbreak\,1.0)$ as the optimal parameter setting and held it fixed in all experiments.

Each EMS-anchor pair forms one record containing up to two orientation-specific actions. The record is assigned the smallest geometric exposure cost $C_{\mathrm{OG}}$ among its eligible actions, and inherits the bottom-support ratio and resting position of that same action, while both actions remain represented if the record is retained. Selection proceeds in staged passes with bottom-support thresholds $0.95$, $0.80$, $0.65$, and a final pass without a support threshold. Within each pass, previously unselected records are sorted by cost and retained when their minimum-cost action meets the current threshold. Near-duplicate records are then suppressed using a bucket with edge length $80\,\mathrm{mm}$ applied to all three resting-position coordinates. Algorithm~\ref{alg:ogems_selection} (in Supplementary Material~\ref{app:ogems_selection}) states the full procedure. The action-level mask acts separately, removing padding rows and rows whose orientation is disallowed or whose placement is geometrically invalid.

Up to $K=64$ region-anchor records are retained. Each retained record contributes up to two orientation-specific action rows, so the number of generated candidate actions satisfies $N_t\leq 2K$. These actions are stored in a fixed $2K$-row candidate table, with unused rows padded and masked. Every downstream policy selects from this same bounded action space.

\subsection{Industrial Candidate Representation}
\label{sec:representation}

Each retained region-anchor-orientation placement is represented as a separate 15-dimensional action row. For $K=64$ retained region-anchor records and up to two allowed orientations per record, the candidate matrix is
\begin{equation}
X_t \in \mathbb{R}^{2K \times 15}.
\end{equation}
The feature vector for candidate action $a$ is
\begin{equation}
\begin{split}
\phi(a)={}&
(x,y,z,d_x,d_y,d_z,f_x,f_y,\\
&\phantom{(}s,m,\hat{h},\ell,b_{\mathrm{frag}},s_{\rm side},\tau),
\end{split}
\end{equation}
where $(x,y,z)$ is the candidate placement anchor; $(d_x,d_y,d_z)$ is the extent of the generating EMS measured from that anchor to its upper corner; $(f_x,f_y)$ is the item footprint under the corresponding orientation; $s$ is the bottom-support ratio; $m$ is the support-region centering margin; $\hat{h}$ is the normalized top height; $\ell$ is the stack-load ratio imposed on items below; $b_{\mathrm{frag}}$ is a binary fragility-feasibility flag; $s_{\rm side}$ is the side-support ratio; and $\tau$ is the placement-effort score. The flag $b_{\mathrm{frag}}$ indicates whether the load-bearing constraint is satisfied, whereas the continuous stack-load ratio $\ell$ quantifies the load imposed on the supporting items relative to their allowable capacity. Exact definitions of all 15 candidate features are provided in Supplementary Material~\ref{app:candidate_features}.

Because orientation is resolved before encoding, the two action rows associated with the same region-anchor pattern may differ in their realized coordinates, EMS extent, footprint, bottom support, and placement-effort score. The representation therefore avoids aliasing between orientation-specific actions. Dimensions 1-8 describe EMS and orientation geometry, while dimensions 9-15 describe the operational quantities computed for each placement.

The feasibility mask is aligned one-to-one with the $2K$ action rows. Masked and padded actions are excluded from sampling and selection rather than being pooled across orientations. The features describe the geometric and operational differences among the remaining actions, and the policy learns to weigh them.

\subsection{Placement Encoder (PE)}

Let $D$ denote the shared embedding dimension used by the candidate, item, and pallet-state representations. The Placement Encoder converts the 15 scalar attributes of each candidate into a fixed sequence of feature tokens. These tokens are projected to a hidden representation, processed by a lightweight xLSTM stack \citep{beck2024xlstm}, normalized, mean-pooled across the feature dimension, and projected to the shared embedding dimension $D$. The encoder therefore produces the vector
\begin{equation}
E_c \in \mathbb{R}^{2K \times D}.
\end{equation}

Treating individual attributes as feature tokens follows tokenization approaches for tabular inputs \citep{gorishniy2021revisiting}. Their order is fixed and grouped into geometry, operational quantities, feasibility, and placement effort. This allows the encoder to model interactions among attributes before candidate actions are compared.

\subsection{Masked Candidate Ranker}

The shared LRAM core receives the candidate-embedding matrix $E_c$, the pallet-state embedding $e_t^{\mathrm{pal}}\in\mathbb{R}^{D}$, and one item token for each allowed orientation. A compact convolutional encoder maps the pallet state $P_t$, represented by its heightmap, to $e_t^{\mathrm{pal}}$, while an item encoder produces orientation-conditioned item tokens from the incoming item's dimensions and attributes. Candidate embeddings are supplied by the PE, or by the corresponding baseline encoder in the ablation experiments.

The LRAM core jointly mixes the pallet-state, item, and candidate tokens. When enabled, a temporal xLSTM carries context between successive packing decisions \citep{beck2024xlstm}. Each candidate row already represents one specific anchor-orientation action. The actor combines its mixed candidate representation with the item token for that orientation and outputs one scalar logit. Thus, the actor produces a separate scalar logit for every anchor-orientation action.

The critic pools the valid shared representations to estimate the state value. All learned policies are trained using proximal policy optimization (PPO) \citep{schulman2017ppo}. Full optimization and implementation details are provided in Supplementary Material~\ref{app:hyperparams}.

\section{Experimental Setup}

\subsection{Dataset and protocol}

We evaluate on 1500 pallet-order instances derived from the BED-BPP benchmark \citep{kagerer2023bedbpp}, a real-world grocery-logistics dataset, using a Euro-pallet footprint. Each order is a fixed item sequence. Dimensions are scaled into the pallet environment, and item metadata (weight, product group, sequence order, rotation allowance) are retained where available. Training convergence is shown in
Supplementary Material~\ref{app:trainingcurves}, and hardware and wall-clock times are reported with the training details in Supplementary Material~\ref{app:hyperparams}. An anonymized implementation is provided in the supplementary material.

\subsection{Compared methods}

OPAL combines OG-EMS, an xLSTM Placement Encoder, and an LRAM backbone without temporal memory. Table~\ref{tab:compared_methods} summarizes the evaluated variants, which examine the effects of temporal memory, the ranking backbone, candidate generation, and the candidate encoder. Greedy OG-EMS provides a non-learned reference using the same candidate generator, while GOPT~\citep{xiong2024gopt}, GENPACK \citep{poolavaram2026genpack}, and PCT \citep{zhao2022pct} serve as external state-of-the-art baselines. We do not evaluate directly against ASAP \citep{fang2025asap} or MPC-3D-BP \citep{fang2026lookahead}, which act at the decision or planning stage and are therefore complementary extensions rather than substitutes. 

\begin{table*}[!h]
\centering
\setlength{\tabcolsep}{3pt}
\renewcommand{\arraystretch}{1.08}
\begin{tabular}{lcccccc}
\toprule
Method
& Abs.\ density
& Rel.\ density
& Surf.\ support
& Side support
& CoG2D
& CoG3D \\
\midrule

OPAL
& $\mathbf{0.49} \pm 0.10$
& $\underline{0.53} \pm 0.12$
& $\underline{0.86} \pm 0.15$
& $0.20 \pm 0.09$
& $\underline{0.84} \pm 0.14$
& $0.57 \pm 0.11$ \\

OPAL w/ temporal
& $\mathbf{0.49} \pm 0.11$
& $0.52 \pm 0.12$
& $0.84 \pm 0.16$
& $0.21 \pm 0.09$
& $\underline{0.84} \pm 0.14$
& $0.58 \pm 0.12$ \\

OPAL w/ Transformer
& $0.42 \pm 0.12$
& $0.45 \pm 0.13$
& $0.67 \pm 0.17$
& $0.22 \pm 0.09$
& $0.77 \pm 0.19$
& $0.50 \pm 0.14$ \\

OPAL w/ Base-EMS
& $\underline{0.48} \pm 0.11$
& $0.52 \pm 0.11$
& $0.85 \pm 0.15$
& $\underline{0.27} \pm 0.11$
& $0.77 \pm 0.14$
& $0.54 \pm 0.10$ \\

OPAL w/ PE=MLP
& $0.45 \pm 0.12$
& $\mathbf{0.55} \pm 0.09$
& $0.84 \pm 0.06$
& $0.25 \pm 0.09$
& $\mathbf{0.91} \pm 0.04$
& $\underline{0.61} \pm 0.04$ \\

\midrule
\multicolumn{7}{l}{\emph{Ablation tier}}\\

Greedy OG-EMS
& $0.46 \pm 0.11$
& $\underline{0.53} \pm 0.15$
& $\mathbf{0.88} \pm 0.16$
& $\mathbf{0.38} \pm 0.12$
& $0.83 \pm 0.16$
& $0.56 \pm 0.13$ \\

\midrule
\multicolumn{7}{l}{\emph{External references}}\\

GOPT (adapted)
& $0.37 \pm 0.12$
& $0.39 \pm 0.12$
& $0.66 \pm 0.15$
& $0.23 \pm 0.08$
& $0.66 \pm 0.16$
& $0.46 \pm 0.10$ \\

PCT
& $0.46 \pm 0.19$
& $0.51 \pm 0.14$
& $0.68 \pm 0.17$
& $0.24 \pm 0.17$
& $0.73 \pm 0.19$
& $0.43 \pm 0.11$ \\

GENPACK
& $0.47 \pm 0.10$
& $\underline{0.53} \pm 0.12$
& $0.81 \pm 0.11$
& $\mathbf{0.38} \pm 0.10$
& $0.80 \pm 0.11$
& $\mathbf{0.62} \pm 0.08$ \\

\bottomrule
\end{tabular}
\caption{Operational KPI means $\pm$ standard deviations over the 1500 reporting orders.}
\label{tab:kpi_mean_std}
\end{table*}

\begin{table*}[!t]
\centering

\begin{minipage}[t]{0.58\textwidth}
\centering

{%
\small
\setlength{\tabcolsep}{2.5pt}
\renewcommand{\arraystretch}{1.05}
\begin{tabular}{lllll}
\toprule
Method & Generator & Encoder & Backbone & Temporal \\
\midrule
OPAL
& OG-EMS & xLSTM & LRAM & off \\

OPAL w/ temporal
& OG-EMS & xLSTM & LRAM & on \\

OPAL w/ Transformer
& OG-EMS & MLP & Transformer & off \\

OPAL w/ Base-EMS
& Base-EMS & xLSTM & LRAM & off \\

OPAL w/ PE=MLP
& OG-EMS & MLP & LRAM & off \\

\midrule
\multicolumn{5}{l}{\emph{Ablation tier}}\\

Greedy OG-EMS
& OG-EMS
& \multicolumn{1}{c}{--}
& \multicolumn{1}{c}{--}
& \multicolumn{1}{c}{--} \\

\midrule
\multicolumn{5}{l}{\emph{External references}}\\

GOPT (adapted)
& Base-EMS
& MLP
& Transformer
& \multicolumn{1}{c}{--} \\

PCT
& \multicolumn{1}{c}{--}
& \multicolumn{1}{c}{--}
& \multicolumn{1}{c}{--}
& \multicolumn{1}{c}{--} \\

GENPACK
& \multicolumn{1}{c}{--}
& \multicolumn{1}{c}{--}
& \multicolumn{1}{c}{--}
& \multicolumn{1}{c}{--} \\
\bottomrule
\end{tabular}
}

\caption{Methods evaluated in this paper.}
\label{tab:compared_methods}
\end{minipage}
\hfill
\begin{minipage}[t]{0.40\textwidth}
\centering

{%
\small
\setlength{\tabcolsep}{3.5pt}
\renewcommand{\arraystretch}{1.05}
\begin{tabular}{lrr}
\toprule
Method
& \multicolumn{1}{c}{Policy}
& \multicolumn{1}{c}{E2E} \\
& \multicolumn{1}{c}{(s/order)}
& \multicolumn{1}{c}{(s/order)} \\
\midrule
OPAL
& $0.38 \pm 0.03$
& $2.88 \pm 1.42$ \\

OPAL w/ temporal
& $0.45 \pm 0.03$
& $3.07 \pm 1.32$ \\

OPAL w/ Transformer
& $\mathbf{0.12} \pm 0.02$
& $2.55 \pm 1.18$ \\

OPAL w/ Base-EMS
& $0.45 \pm 0.03$
& $1.79 \pm 0.80$ \\

OPAL w/ PE=MLP
& $0.27 \pm 0.02$
& $2.64 \pm 1.33$ \\

\midrule
\multicolumn{3}{l}{\emph{Ablation tier}}\\

Greedy OG-EMS
& \multicolumn{1}{c}{--}
& $1.89 \pm 0.97$ \\

\midrule
\multicolumn{3}{l}{\emph{External references}}\\

GOPT (adapted)
& $\underline{0.24} \pm 0.07$
& $\underline{1.16} \pm 0.61$ \\

PCT
& \multicolumn{1}{c}{--}
& $\mathbf{0.25} \pm 0.28$ \\

GENPACK
& \multicolumn{1}{c}{--}
& $32.59 \pm 28.17$ \\
\bottomrule
\end{tabular}
}

\caption{Per-order policy-forward and end-to-end decision-loop latency over 1,500 orders.}
\label{tab:runtime_compact}
\end{minipage}

\end{table*}

\begin{figure*}[!t]
\centering
\includegraphics[width=\textwidth]{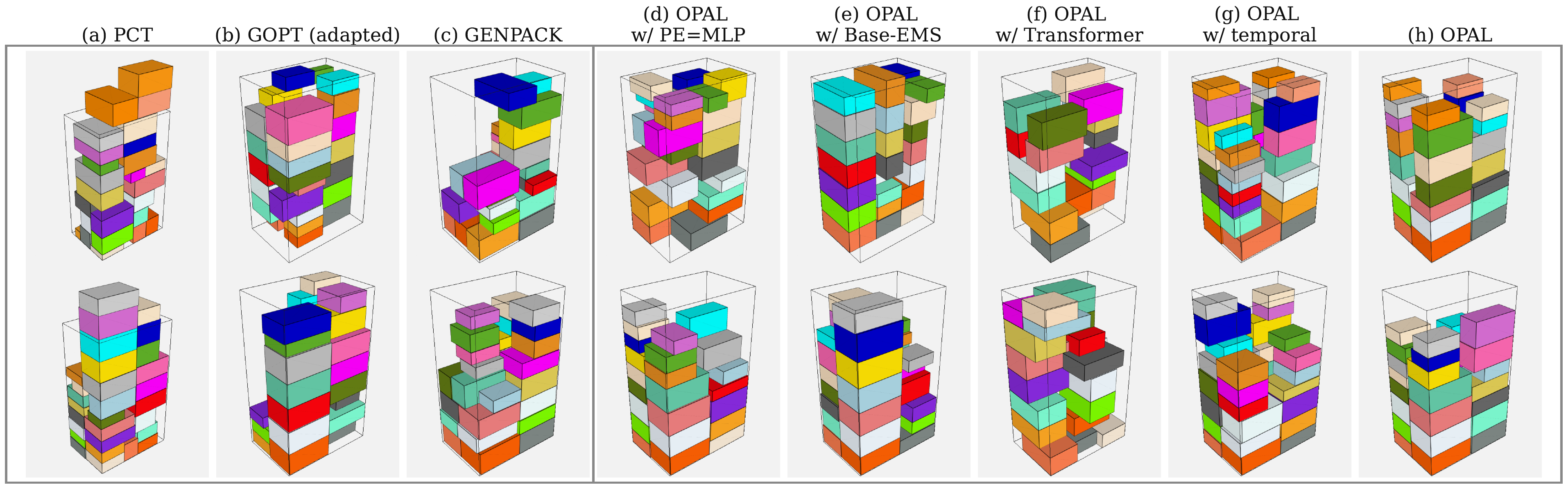}
\caption{Qualitative comparison on two representative orders.}
\label{fig:qualitative}
\end{figure*}

\subsection{Metrics}
\label{sec:metrics}

For every KPI, the raw final-layout score is multiplied by the packing
efficiency $\eta(\mathcal{L})$, the fraction of the order's items placed in the final layout $\mathcal{L}$. For readability, we refer to the resulting reported values by their KPI names throughout the results. This penalizes layouts that reach a high raw score while leaving part of the order unpacked. We additionally report relative density, surface support, side support, and 2D and 3D center-of-gravity (CoG2D and CoG3D, respectively) scores (Table~\ref{tab:kpi_mean_std}). Raw definitions of all KPIs are given in Supplementary Material~\ref{app:kpi_definitions}.

\section{Results}
\label{sec:results}

\subsection{Quantitative results}
\label{sec:ablationsummary}

Table~\ref{tab:kpi_mean_std} summarizes the internal learned configurations together with the non-learned reference and relevant external references. OPAL achieves the highest absolute density across all listed methods. It matches the temporal configuration, and is ahead of the Base-EMS, PE=MLP, and Transformer variants by $0.007$, $0.046$, and $0.070$, respectively. Among the external references, GENPACK is closest in absolute density but remains below OPAL.

The operational KPIs reveal additional trade-offs. OPAL obtains the highest surface support among the learned configurations, while the non-learned Greedy OG-EMS attains the highest surface support overall and ties GENPACK for the highest side support, at markedly lower absolute density. Among the learned configurations, OPAL w/ Base-EMS gives the strongest side support, and PE=MLP records the highest relative density and CoG2D, but at substantially lower absolute density than OPAL.

These comparisons should be interpreted at the configuration level rather than as fully isolated component effects. OPAL and OPAL w/ Base-EMS share the same candidate encoder, backbone, and temporal setting, but differ in both candidate generation and candidate-feature availability. The Transformer configuration also changes the candidate encoder, while PE=MLP replaces the xLSTM Placement Encoder with an MLP.

Greedy OG-EMS reaches an absolute density of $0.46$ over the complete evaluation dataset. OPAL exceeds it by $0.029$, a relative gain of $6.3\%$, measuring the benefit of learned ranking over deterministic selection from the same candidate generator. It further remains competitive on the remaining KPIs, matching OPAL on relative density and trailing it by at most $0.01$ on both center-of-gravity scores, and it attains the highest surface support ($0.88$) and side support ($0.38$) of any listed method. Learned ranking therefore trades support quality for density rather than dominating the deterministic selector outright. This is a learned-versus-heuristic comparison rather than an architectural ablation.

Figure~\ref{fig:qualitative} compares two representative reporting orders. PCT, GOPT, and GENPACK produce more fragmented structures with larger gaps and less regular stacking, whereas the OPAL variants more often form denser lower layers and better-connected stacks. Within the OPAL family, PE=MLP and Transformer show the clearest protrusions and separated columns, while full OPAL remains comparatively compact. These examples are illustrative rather than quantitative.

Two-sided paired Wilcoxon signed-rank tests on matched per-order results \citep{wilcoxon1945individual}, with Holm-Bonferroni correction across the four primary comparisons \citep{holm1979}, show that OPAL is significantly denser than the Transformer, Base-EMS, and PE=MLP variants, but indistinguishable from the temporal variant (Supplementary Material~\ref{app:statistical_significance}).

\subsection{Robustness and deployment characteristics}
\label{sec:robustness}

\paragraph{Pallet-footprint sensitivity.}
We evaluate whether the candidate interface transfers to alternative pallet footprints without architectural changes. Table~\ref{tab:pallet_main} shows that OPAL remains ahead of OPAL w/ Base-EMS at every tested footprint, although the magnitude of the difference is non-monotonic. The largest observed gain occurs at 800$\times$600\,mm, while the difference nearly disappears at
600$\times$400\,mm, where both methods pack only a small fraction of the order volume. The full evaluation methodology is provided in Supplementary Material~\ref{app:pallet}.

\begin{table}[!t]
\centering
\setlength{\tabcolsep}{5pt}
\renewcommand{\arraystretch}{1.05}
\begin{tabular}{lrrr}
\toprule
Footprint (mm) & OPAL & w/ Base-EMS & $\Delta$ \\
\midrule
1200$\times$800  & 0.49 & 0.48 & $+0.007$ \\
1200$\times$1000 & 0.45 & 0.44 & $+0.009$ \\
800$\times$600   & 0.33 & 0.17 & $+0.153$ \\
600$\times$400   & 0.16 & 0.15 & $+0.003$ \\
\bottomrule
\end{tabular}
\caption{Absolute density as a function of pallet footprint, for OPAL and OPAL w/ Base-EMS. }
\label{tab:pallet_main}
\end{table}

\paragraph{Sequence-order robustness.}
The primary experiments present items in descending footprint area. Without
retraining, replacing this ordering with a fixed random permutation reduces
absolute density from 0.491 to 0.445, while reversing the sequence
gives 0.450.
OPAL therefore benefits substantially from the sequencing convention used during training, which deployment should preserve or otherwise retrain for (Supplementary Material~\ref{app:robustness}).

\paragraph{Inference latency.}
Table~\ref{tab:runtime_compact} reports policy-forward and end-to-end decision-loop latency for the internal learned configurations and the external references. Policy-forward latency isolates the neural policy call, whereas end-to-end latency includes candidate generation, masking, action application, and environment bookkeeping. The OPAL family spans $1.79$ to $3.07\,$s per order end-to-end, an order of magnitude below GENPACK, so every variant stays within a range compatible with industrial palletizing cycle times. PCT has the lowest reported end-to-end latency, but runtime should be read jointly with its packing-quality and KPI results.

\section{Discussion}

The KPI profile shows that absolute density alone does not fully characterize industrial packing quality. Relative to OPAL w/ Base-EMS, OPAL achieves higher density, surface support, and center-of-gravity scores, whereas OPAL w/ Base-EMS achieves higher side support. The PE=MLP configuration obtains stronger values on several compactness and balance metrics but substantially lower absolute density. These trade-offs show that no single configuration dominates every operational objective and that deployment decisions should consider the complete KPI profile. Overall, OPAL exceeds GENPACK, a KPI-guided genetic hybrid that optimizes industrial indicators directly, on absolute density (0.49 versus 0.47) and surface support (0.86 versus 0.81) at roughly an order of magnitude lower latency, while GENPACK retains the advantage in side support and CoG3D. OPAL reaches this profile as a single online pass with no post-processing, whereas GENPACK applies a post-processing stage on top of its core method to optimize the final layout. To our knowledge, this is the first learned online policy to surpass a KPI-guided hybrid on absolute density in this industrial setting.

The deterministic generator comparison, which replaces the learned ranker with a shared operational selector so that only the generator differs, provides the clearest evidence for candidate exposure: absolute density increases from 0.40 with Base-EMS to 0.46 with OG-EMS (Supplementary Material~\ref{app:generator_exposure}). Learned comparisons are less isolated because OPAL versus OPAL w/ Base-EMS changes both candidate generation and feature availability, while the Transformer and PE=MLP variants also alter multiple design choices. We therefore interpret these as configuration-level comparisons, with the deterministic experiment serving as the cleaner exposure diagnostic.

The ordering between OPAL and OPAL w/ Base-EMS is preserved across the tested pallet footprints, although the magnitude varies substantially: the interface advantage is largest when the footprint leaves useful placement alternatives to expose and distinguish, and nearly disappears when the action space becomes severely constrained. Heuristic lookahead is similarly non-uniform, improving mean absolute density from 0.46 to 0.48 on the evaluated subset (Supplementary Material~\ref{app:lookahead}) but not benefiting every order. This is consistent with EMS fragmentation, where a locally attractive placement can create free-space regions that are poorly matched to later items, so selective activation may be more appropriate than applying lookahead to every order.

\section{Conclusion}

We presented OPAL, an Operationally Guided Placement-Aware Learning framework for industrial online 3D bin packing. OPAL combines operationally guided candidate exposure, an action-level industrial representation, and masked learned ranking, achieving a mean absolute density of 0.49 across three seeds. Under a common deterministic selector, OG-EMS improves absolute density over Base-EMS by 0.061, a relative gain of $15.1\%$, demonstrating that candidate exposure materially affects packing performance. The learned OPAL configuration also outperforms its Base-EMS counterpart, and this ordering persists across the tested pallet footprints. The operational KPIs reveal meaningful trade-offs among density, support, and balance, but the OPAL family as a whole compares favorably with the contextual references. Future work is targeted to combine OPAL with decision-time mechanisms such as ASAP or MCTS lookahead, and evaluate the resulting packings in physical or high-fidelity robotic settings.

\bibliography{aaai2027}

\clearpage
\appendix

\section{Online Packing Algorithm}
\label{app:algorithm}

Algorithm~\ref{alg:opal} states the decision loop executed for one order. The temporal recurrent state $\mathrm{rec}_t$ is carried across steps only in the temporal configuration.

\begin{algorithm}[!h]
\caption{OPAL online packing loop}
\label{alg:opal}
\begin{algorithmic}[1]
\STATE Initialize the empty pallet state $P_1$.
\STATE Initialize the cross-step temporal state $\mathrm{rec}_0$ when temporal memory is enabled.
\FOR{$t=1$ to $T$}
    \STATE Generate EMS regions from $P_t$.
    \STATE Evaluate the configured anchor patterns and allowed orientations for item $i_t$.
    \STATE Retain up to $K$ region-anchor records (Algorithm~\ref{alg:ogems_selection}).
    \STATE Construct $X_t\in\mathbb{R}^{2K\times15}$ and $M_t\in\{0,1\}^{2K}$.
    \STATE Encode each candidate-action row with the Placement Encoder.
    \STATE Encode the pallet state and orientation-conditioned item inputs.
    \STATE Mix pallet, item, and candidate tokens using the LRAM core.
    \IF{temporal memory is enabled}
        \STATE Update the temporal recurrent state $\mathrm{rec}_t$.
    \ENDIF
    \STATE Produce one scalar logit for each of the $2K$ action rows.
    \STATE Mask disallowed, geometrically invalid, and padded rows using $M_t$.
    \IF{no feasible action remains}
        \STATE Terminate the order.
        \STATE \textbf{break}
    \ENDIF
    \STATE Select action $a_t$, place item $i_t$, and update the pallet state to $P_{t+1}$.
\ENDFOR
\STATE Return the final packed layout.
\end{algorithmic}
\end{algorithm}

\section{OG-EMS Geometric Exposure Cost}
\label{app:ogems_cost}

The geometric exposure quantities are computed in the environment's
discretized length units, where one grid cell corresponds to
$10\,\mathrm{mm}$; the primary $1200\times800\times2000\,\mathrm{mm}$
loading space therefore has dimensions
$L\times W\times H=120\times80\times200$ cells.

Consider an orientation-specific action with footprint anchor $(x,y)$
and oriented footprint dimensions $(f_x,f_y)$, generated from an EMS
with bounds
$[x_{\mathrm{E}}^-,x_{\mathrm{E}}^+]\times
 [y_{\mathrm{E}}^-,y_{\mathrm{E}}^+]\times
 [z_{\mathrm{E}}^-,z_{\mathrm{E}}^+]$.
Only the horizontal bounds enter $C_{\mathrm{OG}}$; the vertical bound
$z_{\mathrm{E}}^+$ is used by the candidate features of Supplementary
Material~\ref{app:candidate_features}.
The gaps from the footprint to the four pallet walls are
\begin{equation}
\begin{aligned}
g_x^- &= x, & g_x^+ &= \max(L-x-f_x,\,0),\\
g_y^- &= y, & g_y^+ &= \max(W-y-f_y,\,0).
\end{aligned}
\end{equation}
The five terms of $C_{\mathrm{OG}}$ are defined below.

\paragraph{Placement height.}
$z$ is the resting height of the footprint, i.e.\ the maximum value of
the pallet heightmap $\mathcal{H}$ beneath it:
\begin{equation}
z=\max_{\substack{0\leq\delta_x<f_x\\0\leq\delta_y<f_y}}
\mathcal{H}(x+\delta_x,\,y+\delta_y).
\end{equation}

\paragraph{Wall distance.}
$d_w$ is the smallest of the four wall gaps:
\begin{equation}
d_w=\min(g_x^-,\,g_x^+,\,g_y^-,\,g_y^+).
\end{equation}

\paragraph{Residual slack.}
Let the item's horizontal clearances inside its EMS be
\begin{equation}
\begin{aligned}
\mathrm{res}_x &= \max(x_{\mathrm{E}}^+-x_{\mathrm{E}}^--f_x,\,0),\\
\mathrm{res}_y &= \max(y_{\mathrm{E}}^+-y_{\mathrm{E}}^--f_y,\,0).
\end{aligned}
\end{equation}
The residual slack is their sum,
$\mathrm{res}=\mathrm{res}_x+\mathrm{res}_y$.

\paragraph{Narrow-slack penalty.}
With a fixed sliver threshold $\varepsilon_{\mathrm{sliv}}=6$ cells
($60\,\mathrm{mm}$),
\begin{equation}
v=
\mathrm{res}_x\,\mathbf{1}[0<\mathrm{res}_x<\varepsilon_{\mathrm{sliv}}]
+
\mathrm{res}_y\,\mathbf{1}[0<\mathrm{res}_y<\varepsilon_{\mathrm{sliv}}],
\end{equation}
so an exact fit or a clearance of at least $\varepsilon_{\mathrm{sliv}}$
contributes nothing on that axis.

\paragraph{Corner distance.}
$d_c$ is the Manhattan distance to the nearest pallet corner, i.e.\ the
smallest sum of one $x$-gap and one $y$-gap:
\begin{equation}
d_c=\min(
g_x^-{+}g_y^-,\,
g_x^-{+}g_y^+,\,
g_x^+{+}g_y^-,\,
g_x^+{+}g_y^+).
\end{equation}

\paragraph{Combined cost.}
The five terms combine linearly,
\begin{equation}
C_{\mathrm{OG}}
=\beta_h z+\beta_w d_w+\beta_{\mathrm{res}}\mathrm{res}+\beta_v v+\beta_c d_c,
\end{equation}
with fixed coefficients
\begin{equation}
(\beta_h,\beta_w,\beta_{\mathrm{res}},\beta_v,\beta_c)
=(5.0,\,2.0,\,0.8,\,3.0,\,1.0).
\end{equation}
All five terms are unnormalized cell lengths, so $C_{\mathrm{OG}}$ is a
nonnegative heuristic cost that is lowest for low, wall- and
corner-adjacent, tightly fitting placements. Each EMS-anchor record
contains up to two eligible orientation-specific actions and is assigned
the minimum $C_{\mathrm{OG}}$ among them.

\section{OG-EMS Candidate Selection}
\label{app:ogems_selection}

Section~\ref{sec:ogems} summarizes candidate selection in prose. This
section states the procedure exactly. Within each EMS region, the five
anchor patterns of Section~\ref{sec:ogems} are instantiated as the
fractional offsets
\begin{equation}
\Xi=
\{(0,0),\,(1,0),\,(0,1),\,(1,1),\,(\tfrac12,\tfrac12)\},
\end{equation}
where an offset $(a_x,a_y)\in\Xi$ places the oriented
footprint at
$x=x_{\mathrm{E}}^-+a_x\max(x_{\mathrm{E}}^+-x_{\mathrm{E}}^--f_x,0)$ and
$y=y_{\mathrm{E}}^-+a_y\max(y_{\mathrm{E}}^+-y_{\mathrm{E}}^--f_y,0)$,
rounded to the nearest cell. The first four are the region's bottom corners
and the fifth is its center. Let $\mathcal{R}_t$ denote the set
of EMS-anchor records generated at step $t$, before any selection is
applied. A record $r\in\mathcal{R}_t$ groups the eligible
orientation-specific actions sharing one EMS region and one anchor
pattern, so its action set $\mathcal{A}(r)$ satisfies
$|\mathcal{A}(r)|\in\{1,2\}$.

\paragraph{Record attributes.}
Each record inherits its attributes from its cheapest eligible action,
\begin{equation}
a^\star(r)=\operatorname*{arg\,min}_{a\in\mathcal{A}(r)}C_{\mathrm{OG}}(a),
\qquad
C(r)=C_{\mathrm{OG}}\bigl(a^\star(r)\bigr).
\end{equation}
The record support $s(r)=s\bigl(a^\star(r)\bigr)$ and the record position
$p(r)=(x,y,z)$ of $a^\star(r)$ are taken from this same action, where $z$
is its resting height and $s$ is the bottom-support ratio of
Supplementary Material~\ref{app:candidate_features}. Both actions of a
retained record are emitted as separate rows, but only $a^\star(r)$
determines whether the record is retained.

\paragraph{Staged ordering.}
Records are arranged into a total order by four successive passes with
support thresholds $\Sigma=(0.95,\,0.80,\,0.65,\,0)$. Pass $k$ appends
every not-yet-ordered record satisfying $s(r)\geq\Sigma_k$, sorted by
increasing $C(r)$. The final pass is unthresholded, so every record is
ordered exactly once. Write the resulting sequence
$\mathcal{Q}_t=(r_1,\ldots,r_{|\mathcal{R}_t|})$.

\paragraph{Spatial-diversity bucket.}
Records are grouped by a discretized cell of side $\varepsilon_{\mathrm{div}}=8$ cells
($80\,\mathrm{mm}$) applied to all three coordinates of $p(r)$,
\begin{equation}
\kappa(r)=\Bigl(\lfloor x/\varepsilon_{\mathrm{div}}\rfloor,\;
\lfloor y/\varepsilon_{\mathrm{div}}\rfloor,\;
\lfloor z/\varepsilon_{\mathrm{div}}\rfloor\Bigr).
\end{equation}
The vertical coordinate participates, so two records that share a
footprint but rest at different heights occupy different buckets.

\paragraph{Selection.}
Algorithm~\ref{alg:ogems_selection} fills the $K$ record slots in three
stages. A quality stage takes the $K_q=\max(1,\lfloor 0.75K\rfloor)$
leading records of $\mathcal{Q}_t$ with no diversity constraint. A diversity
stage then admits further records only when their bucket is unoccupied.
A backfill stage lifts the bucket constraint if fewer than $K$ records
have been selected. With $K=64$ this reserves $K_q=48$ slots for
cost-ranked quality and up to $16$ slots for spatial spread, while
guaranteeing that the table is filled whenever
$|\mathcal{R}_t|\geq K$. Each retained record then emits its eligible
action rows, giving $N_t\leq 2K$.

\begin{algorithm}[!h]
\caption{OG-EMS record selection at step $t$}
\label{alg:ogems_selection}
\begin{algorithmic}[1]
\STATE \textbf{Input:} records $\mathcal{R}_t$, budget $K$, quota $K_q$, cell $\varepsilon_{\mathrm{div}}$.
\STATE Order $\mathcal{R}_t$ into $\mathcal{Q}_t$ by the staged passes over $\Sigma$.
\STATE Initialize the ordered selection list $\mathcal{D}\leftarrow()$.
\FOR{$r$ in $\mathcal{Q}_t$}
    \STATE Append $r$ to $\mathcal{D}$.
    \IF{$|\mathcal{D}|\geq K_q$}
        \STATE \textbf{break}
    \ENDIF
\ENDFOR
\STATE Initialize the occupied buckets $\mathcal{B}\leftarrow\{\kappa(r):r\in \mathcal{D}\}$.
\FOR{$r$ in $\mathcal{Q}_t$ not already in $\mathcal{D}$, while $|\mathcal{D}|<K$}
    \IF{$\kappa(r)\notin \mathcal{B}$}
        \STATE Append $r$ to $\mathcal{D}$ and set $\mathcal{B}\leftarrow \mathcal{B}\cup\{\kappa(r)\}$.
    \ENDIF
\ENDFOR
\FOR{$r$ in $\mathcal{Q}_t$ not already in $\mathcal{D}$, while $|\mathcal{D}|<K$}
    \STATE Append $r$ to $\mathcal{D}$.
\ENDFOR
\STATE Return $\mathcal{D}$, containing at most $K$ records in selection order.
\end{algorithmic}
\end{algorithm}

\section{Candidate-Feature Definitions}
\label{app:candidate_features}

Let $(L,W,H)$ denote the pallet loading dimensions, where $H$ is the loading
height; let $(x,y,z)$ be the lower placement corner; let
$(u_x,u_y,u_z)=(x_{\mathrm{E}}^+,y_{\mathrm{E}}^+,z_{\mathrm{E}}^+)$ be the
upper corner of the generating EMS in the notation of Supplementary
Material~\ref{app:ogems_cost}; and let $(f_x,f_y,h)$ be the oriented item
dimensions, where $h$ denotes the item height. Let
\begin{equation}
\Omega(f_x,f_y)
=
\{0,\ldots,f_x-1\}
\times
\{0,\ldots,f_y-1\}
\end{equation}
denote the set of discrete footprint-cell offsets for an item with oriented
footprint dimensions $(f_x,f_y)$. Table~\ref{tab:candidate_features} defines
the 15 entries of each action row exactly as supplied to the candidate
encoder. Entries 1--6 describe the record's anchor geometry and are inherited
from $a^\star(r)$ as defined in Supplementary
Material~\ref{app:ogems_selection}, so both action rows of a record share
them. Entries 7--15 are orientation-specific: the footprint is that of the
row's own orientation, and every operational quantity is evaluated at that
orientation's placement. The first eight entries remain in the environment's
discretized length units; they are not divided by the pallet dimensions. The
operational ratios are dimensionless, top height is normalized by the pallet
loading height, and the placement-effort score is bounded below by its
constant term $\lambda_0=1.0$, defined in Supplementary
Material~\ref{app:reward_definitions}.

\begin{table*}[!t]
\centering

\setlength{\tabcolsep}{4pt}
\renewcommand{\arraystretch}{1.12}
\begin{tabular}{ccll}
\toprule
Index & Feature & Exact definition & Range \\
\midrule
1 & $x$
& Lower placement coordinate on the first pallet axis
& $[0,L-f_x]$ \\

2 & $y$
& Lower placement coordinate on the second pallet axis
& $[0,W-f_y]$ \\

3 & $z$
& Lower placement height
& $[0,H-h]$ \\

4 & $d_x$
& $u_x-x$, available EMS extent from the anchor to its upper boundary
& $[0,L]$ \\

5 & $d_y$
& $u_y-y$, available EMS extent from the anchor to its upper boundary
& $[0,W]$ \\

6 & $d_z$
& $u_z-z$, available EMS extent from the anchor to its upper boundary
& $[0,H]$ \\

7 & $f_x$
& Oriented item footprint on the first pallet axis
& $(0,L]$ \\

8 & $f_y$
& Oriented item footprint on the second pallet axis
& $(0,W]$ \\

9 & $s$
& $\displaystyle
\frac{
\left|
\left\{
(\delta_x,\delta_y)\in\Omega(f_x,f_y):
\mathcal{H}(x+\delta_x,y+\delta_y)=z
\right\}
\right|
}{
f_xf_y
}$;
$s=1$ on the pallet floor
& $[0,1]$ \\

10 & $m$
& $0.5$ if $z=0$ and $0$ if no support cell exists
& \\
& &
otherwise
$\displaystyle
\max\!\left\{
0,\,
\frac{
\min(f_x/2-\delta_x^{\min},\,
     \delta_x^{\max}-f_x/2,\,
     f_y/2-\delta_y^{\min},\,
     \delta_y^{\max}-f_y/2)
}{
\max(f_x,f_y)
}
\right\}$
& $[0,0.5]$ \\

11 & $\hat h$
& $(z+h)/H$, normalized top height
& $[0,1]$ \\

12 & $\ell$
& $\displaystyle\max_{j\in\mathcal{S}}\omega_i/\bar\omega_j$;
  $\ell=0$ when $\mathcal{S}=\varnothing$
& $[0,\infty)$ \\

13 & $b_{\mathrm{frag}}$
& $\mathbf{1}[\ell\leq1]$, fragility-feasibility indicator
& $\{0,1\}$ \\

14 & $s_{\rm side}$
& Supported non-boundary side faces divided by non-boundary side faces
& $[0,1]$ \\

15 & $\tau$ & Placement-effort score & $[\lambda_0,\infty)$ \\
\bottomrule
\end{tabular}

\caption{Candidate-action features and their ranges for generated actions.
$\mathcal{H}$ denotes the heightmap beneath the oriented footprint;
$[\delta_x^{\min},\delta_x^{\max}) \times [\delta_y^{\min},\delta_y^{\max})$ is the half-open
axis-aligned bounding box of footprint-cell offsets supported at height $z$, so
$\delta_x^{\max}$ is one past the largest supported offset on that axis;
$\mathcal{S}$ is the set of items directly supporting the candidate;
$\omega_i$ is the candidate-item weight; and $\bar\omega_j$ is the load
capacity of supporting item $j$. The first eight features use the
environment's discretized length units, whereas $\hat h$ is normalized
by the pallet loading height and $\tau$ is a placement-effort score bounded
below by its constant term $\lambda_0=1.0$.}
\label{tab:candidate_features}
\end{table*}

For $s_{\rm side}$, pallet-wall faces are excluded. Each remaining side face is marked supported when item-to-item contact covers at least $20\%$ of its area. The candidate-level placement-effort score $\tau$ uses the definition and coefficients given in Supplementary Material~\ref{app:reward_definitions}, evaluated for the candidate placement rather than for the committed placement.

\section{Reward-Signal Definitions}
\label{app:reward_definitions}

Let $f_t$ denote the raw absolute density after decision step $t$, and let
$\Delta f_t=f_t-f_{t-1}$. For a successful placement of item $i_t$ with height
$h_t$ at vertical coordinate $z_t$, its normalized top
height is
\begin{equation}
\hat h_t=\frac{z_t+h_t}{H}.
\end{equation}

\paragraph{Incremental raw absolute density.}
\begin{equation}
\Delta f_t=\frac{V_{i_t}}{LWH},
\end{equation}
where $V_{i_t}$ is the volume of the placed item $i_t$.

\paragraph{Bottom support.}
$S_t$ is the fraction of footprint cells whose height equals the
placement height. Ground placements receive $S_t=1$.

\paragraph{Support-region centering.}
$B_t\in[0,0.5]$ is the minimum distance from the center of the item footprint to the four edges of the axis-aligned bounding box of the supported footprint cells, normalized by $\max(f_x,f_y)$. Ground placements receive $B_t=0.5$, while placements with no supported footprint cells receive $B_t=0$. It is the reward-level counterpart of the candidate feature $m$, and it is a purely geometric measure that does not use item mass.

\paragraph{Placement effort.}
The placement-effort score combines vertical placement height,
orientation change, reachability, and insufficient bottom support:
\begin{equation}
\begin{split}
\tau_t ={}&
\lambda_0
+\lambda_h(z_t+h_t)
+\mathbf{1}_{\rm rot}\lambda_{\rm rot} \\
&{}+
\mathbf{1}_{\rm unreachable}\lambda_{\rm reach}
+\lambda_s\max(0,0.75-S_t),
\end{split}
\end{equation}
where $\mathbf{1}_{\rm rot}$ indicates whether the placement requires
rotation and
$\mathbf{1}_{\rm unreachable}=\mathbf{1}[z_t+h_t+\varepsilon_{\mathrm{grip}}
>H_{\mathrm{reach}}]$ indicates that the placement violates the configured
reach condition of Supplementary Material~\ref{app:hyperparams}. The fixed
coefficients are
\begin{equation}
\begin{aligned}
\lambda_0 &= 1.0, &
\lambda_h &= 0.004, &
\lambda_{\rm rot} &= 0.25,\\
\lambda_{\rm reach} &= 0.4, &
\lambda_s &= 0.5.
\end{aligned}
\end{equation}
Higher values indicate placements associated with greater operational effort.

\paragraph{Instability risk.}
Let $\ell_t$ be the maximum stack-load ratio imposed on directly supporting items. Then
\begin{equation}
R_t=(1-S_t)+\max(0,\ell_t-1).
\end{equation}

\paragraph{Height terms.}
Let $z_t^{\max}$ denote the pallet maximum height after step $t$. The fill-weighted placement-height and height-growth penalties are
\begin{align}
\chi_t^{\mathrm{height}}
&= \hat h_t\Delta f_t,\\
G_t
&=
\frac{\max(0,z_t^{\max}-z_{t-1}^{\max})}{H}
\Delta f_t.
\end{align}

\paragraph{Low-placement reward.}
\begin{equation}
U_t=\Delta f_t(1-\hat h_t).
\end{equation}
Note that $U_t+\chi_t^{\mathrm{height}}=\Delta f_t$, so $U_t$ and
$\chi_t^{\mathrm{height}}$ are collinear given $\Delta f_t$; the pair
expresses one height preference through two weights rather than two
independent signals.

\paragraph{Wall proximity.}
Let $d_t^{\mathrm{wall}}$ be the minimum distance from the placed footprint to the nearest pallet boundary, and let
$d_{\max}=\min(L,W)/2$. Then
\begin{equation}
\chi_t^{\mathrm{wall}}
=
\Delta f_t
\max\left(
0,
1-\frac{d_t^{\mathrm{wall}}}{d_{\max}}
\right).
\end{equation}

\paragraph{Lateral support.}
$\chi_t^{\mathrm{side}}$ is the fraction of non-boundary side faces of the placed item whose contact with previously placed items covers at least $20\%$ of the face area.
\paragraph{Terminal transitions.}
A failed terminal transition receives a penalty of $-w_{\mathrm{fail}}$, where $w_{\mathrm{fail}}=0.5$. At any terminal transition, a terminal fill term $0.6f_T$ is added, where $f_T$ is the final raw absolute density. Thus, a successful completion receives the
final placement reward plus $0.6f_T$, whereas an unsuccessful termination receives
\begin{equation}
-0.5+0.6f_T.
\end{equation}

\section{Operational KPI Definitions}
\label{app:kpi_definitions}

Let $\mathcal{L}$ denote the final packed layout, $\mathcal{I}$ the complete set of items in the order, and $\mathcal{P}\subseteq\mathcal{I}$ the items retained in $\mathcal{L}$. For item $i\in\mathcal{P}$, let $(x_i,y_i,z_i)$ denote its lower placement corner and $(w_i,l_i,h_i)$ its placed dimensions, where $w_i$, $l_i$, and $h_i$ are its extents along the first, second, and vertical pallet axes respectively, so that $w_i$ runs parallel to $L$ and $l_i$ parallel to $W$; its volume is $V_i=w_i l_i h_i$.

The packing efficiency is
\begin{equation}
\eta(\mathcal{L})
=
\frac{|\mathcal{P}|}{|\mathcal{I}|}.
\end{equation}
When $\mathcal{P}=\varnothing$, every layout-level KPI score is defined to be $0$.
The definitions below give the raw geometric and operational KPIs. Each value
reported in the experiments is the corresponding normalized score
\begin{equation}
\widetilde{\operatorname{KPI}}(\mathcal{L})
=
\eta(\mathcal{L})\,\operatorname{KPI}(\mathcal{L}),
\label{eq:kpi_normalization}
\end{equation}
which penalizes layouts that reach a high raw score while leaving part of the
order unpacked. Following the main paper, reported KPI values are given to two
decimal places, except where finer resolution is needed to separate nearby
values, and differences between them to three. All differences are computed
from the unrounded scores, so they need not equal the difference of the
rounded values shown.

\paragraph{Absolute density.}
The raw absolute density, equivalently the space utilization, is the fraction of the pallet loading volume occupied by placed items \citep{martello2000three,crainic2008extreme}:
\begin{equation}
\operatorname{AbsDen}(\mathcal{L})=\frac{\sum_{i\in\mathcal{P}}V_i}{LWH}.
\end{equation}
The denominator uses the configured loading height $H$, not the maximum height reached by the packing. 

\paragraph{Relative density.}
Relative density measures the occupied fraction of the smallest
axis-aligned bounding box enclosing the placed items. Let
\begin{align}
x^- &= \min_{i\in\mathcal{P}} x_i,
& x^+ &= \max_{i\in\mathcal{P}} (x_i+w_i),\\
y^- &= \min_{i\in\mathcal{P}} y_i,
& y^+ &= \max_{i\in\mathcal{P}} (y_i+l_i),\\
z^- &= \min_{i\in\mathcal{P}} z_i,
& z^+ &= \max_{i\in\mathcal{P}} (z_i+h_i).
\end{align}
The relative-density score is
\begin{equation}
\operatorname{RelDen}(\mathcal{L})
=
\frac{\sum_{i\in\mathcal{P}} V_i}
{(x^+-x^-)(y^+-y^-)(z^+-z^-)}.
\end{equation}
The score is clipped to $[0,1]$. Higher values indicate a more compact arrangement within the enclosing bounding box.

\paragraph{Surface support.}
For item $i$, let $c_i^{\rm surf}\in[0,1]$ be the fraction of its bottom face covered by items whose top face is at $z_i$, and let $n_i^{\rm corner}$ count the bottom-corner support incidences at that height, so that a corner covered by two items contributes twice. Its surface-support score, clipped to $[0,1]$, is
\begin{equation}
\sigma_i=
\begin{cases}
1,
& z_i=0,\\
1,
& c_i^{\rm surf}\geq0.5 \ \text{and}\ n_i^{\rm corner}\geq3,\\
n_i^{\rm corner}/4,
& c_i^{\rm surf}<0.5 \ \text{and}\ n_i^{\rm corner}\geq3,\\
c_i^{\rm surf},
& \text{otherwise}.
\end{cases}
\end{equation}
The raw surface-support score of the final layout is
\begin{equation}
\operatorname{SurfSup}(\mathcal{L})
=
\frac{1}{|\mathcal{P}|}
\sum_{i\in\mathcal{P}}\sigma_i.
\end{equation}

\paragraph{Side support.}
For item $i$, let $\mathcal{F}^{\rm side}_i$ denote the side faces that do not coincide with a pallet boundary. For each $g\in\mathcal{F}^{\rm side}_i$, let $A_{i,g}$ be its area and $A^{\rm con}_{i,g}$ the area in contact with neighboring items. A face is considered supported when at least $20\%$ of its area is in contact with another item, consistent with common industrial loading requirements \citep{bortfeldt2013constraints}. The layout-level
side-support score is
\begin{equation}
\operatorname{SideSup}(\mathcal{L})
=
\frac{
\displaystyle
\sum_{i\in\mathcal{P}}
\sum_{g\in\mathcal{F}^{\rm side}_i}
\mathbf{1}\!\left[
\frac{A^{\rm con}_{i,g}}{A_{i,g}}\geq0.2
\right]
}{
\displaystyle
\sum_{i\in\mathcal{P}}
\left|\mathcal{F}^{\rm side}_i\right|
}.
\end{equation}
The score is defined to be $0$ when the denominator vanishes, that is, when
every side face of every placed item coincides with a pallet boundary.

\paragraph{Center of gravity.}
Let $\mu_i$ denote the item mass, computed from reported density times volume when density is available, from reported weight otherwise, and from item volume when neither is given. The mass $\mu_i$ is distinct from the item weight $\omega_i$ of Supplementary Material~\ref{app:candidate_features}, which enters the stack-load ratio. Because $\mu_i$ appears in both the numerator and the denominator below, any constant unit factor between the two cancels. The mass-weighted planar center of gravity is
\begin{equation}
(x_c^{(2)},y_c^{(2)})=
\frac{
\sum_{i\in\mathcal{P}}\mu_i
\left(x_i+\frac{w_i}{2},y_i+\frac{l_i}{2}\right)}
{\sum_{i\in\mathcal{P}}\mu_i}.
\end{equation}
The mass-weighted three-dimensional center of gravity is
\begin{equation}
(x_c^{(3)},y_c^{(3)},z_c^{(3)})=
\frac{
\sum_{i\in\mathcal{P}}\mu_i
\left(x_i+\frac{w_i}{2},y_i+\frac{l_i}{2},
z_i+\frac{h_i}{2}\right)}
{\sum_{i\in\mathcal{P}}\mu_i}.
\end{equation}

Let
\begin{align}
\rho_2 &=
\frac{\sqrt{(x_c^{(2)}-L/2)^2+(y_c^{(2)}-W/2)^2}}
{\sqrt{(L/2)^2+(W/2)^2}},\\
\rho_3 &=
\frac{\sqrt{(x_c^{(3)}-L/2)^2+(y_c^{(3)}-W/2)^2
+(z_c^{(3)})^2}}
{\sqrt{(L/2)^2+(W/2)^2+H^2}}.
\end{align}
The raw center-of-gravity scores are
\begin{align}
\operatorname{CoG2D}(\mathcal{L})&=[1-\rho_2]_{0}^{1},\\
\operatorname{CoG3D}(\mathcal{L})&=[1-\rho_3]_{0}^{1},
\end{align}
where $[a]_0^1=\min(1,\max(0,a))$ denotes clipping to $[0,1]$. Higher CoG2D values indicate that the projected center of gravity is closer to the center of the pallet footprint. Higher CoG3D values indicate that the center of gravity is closer to the center of the pallet floor, thereby favoring both horizontal balance and a low vertical center of gravity.

\section{Training Convergence}
\label{app:trainingcurves}

\begin{figure}[!t]
\centering
\includegraphics[width=\linewidth]{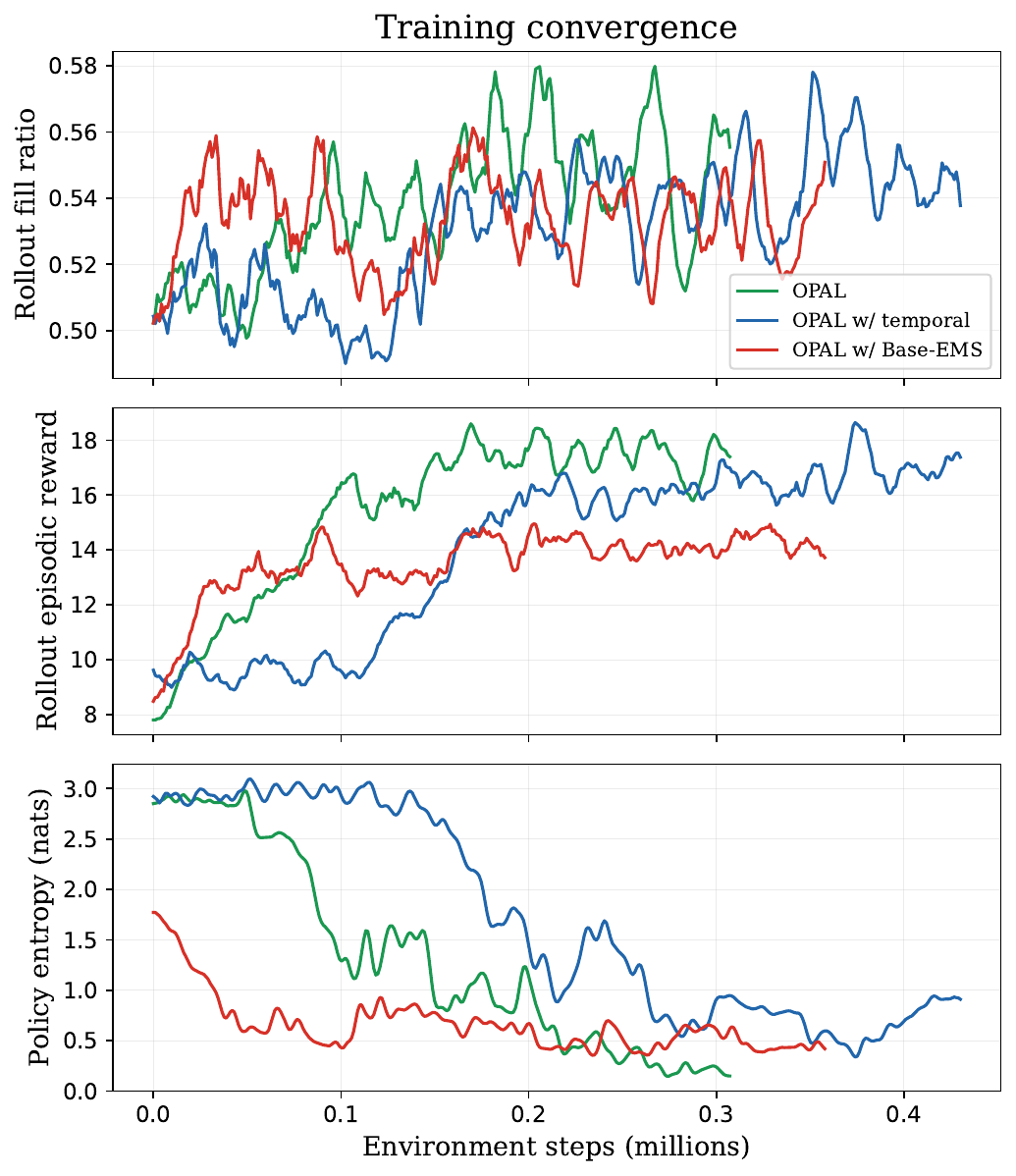}
\caption{Training convergence for one selected run of OPAL, OPAL w/ temporal, and OPAL w/ Base-EMS: raw training-episode fill ratio (top), shaped episodic reward (middle), and policy entropy over the masked
candidate set (bottom). For each variant, the run with the highest peak training-pool evaluation reward among seeds 5--7 is shown; reporting orders are not used for this selection. Curves are smoothed with
centred moving averages, and each line ends when the corresponding run early-stops. Quantitative comparisons use the three-seed held-out evaluation reported in the main paper.}
\label{fig:training_curves}
\end{figure}

Figure~\ref{fig:training_curves} illustrates rollout-based training dynamics for OPAL, OPAL w/ temporal, and OPAL w/ Base-EMS. For visualization, we show the run with the highest peak training-pool evaluation reward among seeds 5--7 for each variant. This gives seed 6 for OPAL and seed 7 for the other two variants. The 1500 reporting orders play no part in this selection. The figure provides a qualitative view of optimization behaviour; all quantitative comparisons use the three-seed held-out evaluation reported in
Table~\ref{tab:kpi_mean_std}.

Each line ends when the corresponding run early-stops; no values are extrapolated. The top panel shows raw training-episode fill ratio, the middle panel shows training episodic reward under the shaped objective of
Supplementary Material~\ref{app:reward_definitions}, and the bottom panel shows the entropy of the policy over the masked candidate set. The training-episode fill ratio is measured under the exploring policy
and is distinct from the packing-efficiency-normalized score reported in the main results.

The fill-ratio trajectories are noisy and overlap substantially. In the displayed runs, OPAL reaches higher episodic rewards earlier, OPAL w/ temporal improves more gradually, and OPAL w/ Base-EMS reaches a lower reward range. Policy entropy decreases during all three displayed runs, consistent with increasing concentration of the masked action distribution. Because the attainable entropy depends on the number of currently unmasked actions, and this number differs between generators, absolute entropy levels should not be compared between OG-EMS and Base-EMS.

\section{Full Hyperparameters and Training Details}
\label{app:hyperparams}

\paragraph{PPO configuration.} The shared embedding dimension is $D=64$. The
xLSTM stack used by the Placement Encoder comprises two blocks with two heads
and hidden size 32, while the LRAM core uses an xLSTM stack of three blocks
with four heads and hidden size 64. In the temporal configuration the temporal
xLSTM adds two blocks with four heads, hidden size 64, and context length 32.
Training uses Adam with learning rate $7.5{\times}10^{-5}$ and no decay, clip range 0.15, discount $\gamma=0.99$, $\lambda_{\mathrm{GAE}}=0.95$, eight parallel environments, rollout length 32 steps per environment per update, and one gradient epoch per update. The entropy coefficient is $7{\times}10^{-4}$, the value-loss coefficient is 0.35, and the maximum gradient norm is 0.35; value clipping, advantage normalization, and reward normalization are enabled. Batch size is 32 for all the training variants. Each run has a nominal ceiling of $10000$ environment steps per training cycle over at most $1000$ training cycles, with early stopping after 10 training-pool evaluations without an improvement of at least $10^{-4}$. Realized length varies by seed and method. Training uses one NVIDIA L40S GPU; OPAL wall-clock times are 4.3, 7.5, and 3.5 hours for seeds 5, 6, and 7, respectively. Every internal learned configuration is evaluated with seeds 5--7.

\paragraph{Feasibility and operational thresholds.}
An action row is admissible when the oriented item fits inside its EMS and the
pallet, and when it satisfies the geometric stability rule of the underlying
masked-selection environment, which accepts a placement if the item rests on
the pallet floor, if more than half of its bottom face is supported, or if the
footprint center lies inside the convex hull of the supported cells
\citep{xiong2024gopt}. This rule, together with containment, defines the mask
$M_t$ of Section~\ref{sec:problemformulation} in all reported
configurations.

The operational quantities that industrial deployment additionally cares
about, namely support, overhang, load bearing, and reach
\citep{bortfeldt2013constraints}, enter the pipeline in two other places
rather than as hard admissibility gates. They are computed for every generated
candidate and supplied to the policy as the features of Supplementary
Material~\ref{app:candidate_features}, and the bottom-support ratio drives the
staged ordering of Supplementary Material~\ref{app:ogems_selection}, whose
thresholds $\Sigma=(0.95,\,0.80,\,0.65,\,0)$ act as retention preferences
under the record budget $K$. The reference thresholds used when reporting
these quantities are a bottom-support ratio $s\geq0.65$, equivalently an
overhang of at most $0.35$; a support-region centering margin $m\geq0.08$, used as an additional
quantitative centering diagnostic alongside the binary convex-hull rule; a stack-load ratio
$\ell\leq1$, so that no item exceeds the rated capacity of the items beneath
it; and a reach condition
$z+h+\varepsilon_{\mathrm{grip}}\leq H_{\mathrm{reach}}$, where
$\varepsilon_{\mathrm{grip}}$ is the headroom the end effector needs above the
placed item and $H_{\mathrm{reach}}$ is the highest point the robot can
service. The last two describe the deployment cell rather than the packing
method, and are configuration inputs rather than tuned quantities; we use
$\varepsilon_{\mathrm{grip}}=4$ cells ($40\,\mathrm{mm}$) and
$H_{\mathrm{reach}}=210$ cells ($2100\,\mathrm{mm}$), which exceeds the
$2000\,\mathrm{mm}$ loading height, so the reach condition is never binding in
the primary configuration and the reported results do not depend on a
particular reach envelope. The environment implements these thresholds as a
composite gate that can be enabled to mask actions directly; it is inactive in
the configurations reported here. All of these values were fixed before
evaluation on the reporting orders and were not tuned. Candidate generation
additionally caps the raw EMS enumeration at 512 regions before anchor
evaluation.

\paragraph{Deterministic operational selector.}
Greedy OG-EMS and the heuristic-lookahead experiments select among the generated candidate actions using
\begin{equation}
\begin{split}
J_{\mathrm{OG}}(P_t,a)={}&
\alpha_s s+
\alpha_{ss}s_{\mathrm{side}}+
\alpha_m m+
\alpha_q q+
\alpha_w \psi_w \\
&{}-
\alpha_h\hat h-
\alpha_o o-
\alpha_\ell\ell-
\alpha_\tau\tau,
\end{split}
\label{eq:operational_selector}
\end{equation}
where $P_t$ is the current pallet state, $s$ is bottom support, $s_{\mathrm{side}}$ is side support, $m$ is the support-region centering margin, $q$ is item volume divided by pallet volume, $\psi_w=1-d_w/d_{\max}$ is the wall-proximity score, where $d_w$ is the nearest-wall distance defined in Supplementary Material~\ref{app:ogems_cost} and $d_{\max}=\min(L,W)/2$, $\hat h$ is normalized top height, $o=1-s$ is overhang,
$\ell$ is the stack-load ratio, and $\tau$ is the placement-effort score.
The default coefficients are
\begin{equation}
\begin{split}
&(\alpha_s,\alpha_{ss},\alpha_m,\alpha_q,\alpha_w,
\alpha_h,\alpha_o,\alpha_\ell,\alpha_\tau)\\
&\qquad=(6.0,\allowbreak\,1.2,\allowbreak\,2.0,\allowbreak\,0.8,
\allowbreak\,0.35,\\
&\qquad\phantom{=(}3.0,\allowbreak\,2.5,\allowbreak\,1.5,
\allowbreak\,0.25).
\end{split}
\end{equation}
Because $q$ is constant across candidates for the current item, it does not affect their within-step ordering. Because $o=1-s$, the support and overhang terms are collinear and act as the single effective support weight $\alpha_s+\alpha_o=8.5$, so varying $\alpha_s$ alone varies this combined weight. This selector is used only by the deterministic baseline and heuristic lookahead; the learned OPAL policy does not use $J_{\mathrm{OG}}$.

\paragraph{OG-EMS weight selection.}
The default selector weights, including $(\alpha_s,\alpha_w)=(6.00,0.35)$, were fixed before the post-hoc sensitivity analysis in Table~\ref{tab:weight_sensitivity}. That analysis is evaluated on 1000 orders drawn from the training pool, which is disjoint from the 1500 reporting orders, so no reporting-set result informed the choice of weights. The sweep is therefore diagnostic rather than a model-selection experiment: the two best settings, $(6.00,1.00)$ and $(6.00,2.00)$, are indistinguishable at $0.47$ absolute density, and no setting separates from the default by a margin that would justify changing it.

\begin{table}[h]
\centering
\small
\setlength{\tabcolsep}{3pt}
\begin{tabular}{ccccc}
\toprule
$\alpha_s$ & $\alpha_w$ & Abs.\ dens. & Surf.\ sup. & Side sup. \\
\midrule
6.00 & 0.35 & $0.46 \pm 0.12$ & $0.87 \pm 0.16$ & $0.37 \pm 0.12$ \\
6.00 & 1.00 & $0.47 \pm 0.11$ & $0.88 \pm 0.16$ & $0.34 \pm 0.12$ \\
4.00 & 1.00 & $0.46 \pm 0.12$ & $0.87 \pm 0.16$ & $0.35 \pm 0.12$ \\
6.00 & 2.00 & $0.47 \pm 0.11$ & $0.88 \pm 0.16$ & $0.32 \pm 0.13$ \\
\bottomrule
\end{tabular}
\caption{Post-hoc sensitivity of the deterministic OG-EMS selector over 1000 evaluation runs drawn from the training pool, disjoint from the reporting orders, reported as normalized scores. The default $(\alpha_s,\alpha_w)=(6.00,0.35)$ was fixed before this analysis and is retained to avoid tuning on the reporting set.}
\label{tab:weight_sensitivity}
\end{table}

\section{Statistical Significance}
\label{app:statistical_significance}

For the four primary learned comparisons, we apply two-sided paired Wilcoxon signed-rank tests to matched per-order absolute density and control the family-wise error rate using Holm-Bonferroni correction. For each method, the three seed results are first averaged per order; the paired tests are then applied to the resulting matched order-level scores. Table~\ref{tab:stat_tests_norm_abs_density} reports the corrected $p$-values.

\begin{table}[h]
\centering
\small
\begin{tabular}{lrrr}
\toprule
Comparison & $\Delta$ Abs.\ dens. & Holm $p$ & Signif. \\
\midrule
OPAL w/ temporal    & $+0.002$ & 0.256 & no \\
OPAL w/ Transformer & $+0.070$ & $<0.001$ & yes \\
OPAL w/ Base-EMS    & $+0.007$ & 0.005 & yes \\
OPAL w/ PE=MLP      & $+0.046$ & $<0.001$ & yes \\
\bottomrule
\end{tabular}
\caption{Paired absolute-density comparisons against OPAL using two-sided Wilcoxon signed-rank tests with Holm-Bonferroni correction across the four comparisons. Positive differences indicate higher absolute density for OPAL.}
\label{tab:stat_tests_norm_abs_density}
\end{table}

\section{Deterministic Generator-Exposure Analysis}
\label{app:generator_exposure}

A candidate generator can improve a learned online packing system only if it exposes useful feasible placements to the selector. Directly computing the best achievable sequential packing under each generator is intractable because each selected placement changes the subsequent packing state and candidate set. We therefore apply a common deterministic selector as a proxy for candidate exposure. This diagnostic removes the learned ranker and evaluates both generators using the same operational scoring rule.

For generator $\mathcal{G}$, let $\mathcal{A}_{\mathcal{G}}(P_t)$ denote the candidate actions produced from pallet state $P_t$. At each step, the deterministic selector chooses
\begin{equation}
a_t
=
\arg\max_{a\in\mathcal{A}_\mathcal{G}(P_t)}
J_{\mathrm{OG}}(P_t,a),
\end{equation}
where $J_{\mathrm{OG}}$ is the operational score of Eq.~\ref{eq:operational_selector}. Its coefficients are manually specified rather than learned and are held fixed across generators.

All operational quantities required by $J_{\mathrm{OG}}$ are computed identically after candidate generation for both Base-EMS and OG-EMS candidates. The geometry-only candidate representation supplied to the learned Base-EMS policy therefore does not constrain this deterministic comparison. Table~\ref{tab:generator_exposure} reports the outcome.

\begin{table}[h]
\centering
\small
\setlength{\tabcolsep}{6pt}
\renewcommand{\arraystretch}{1.05}
\begin{tabular}{lcc}
\toprule
Generator & Greedy norm.\ abs.\ density & OG-EMS $-$ row \\
\midrule
Base-EMS & 0.40 & $+0.061$ \\
OG-EMS & 0.46 & $0.000$ \\
\bottomrule
\end{tabular}
\caption{Deterministic generator comparison over 1500 evaluation runs spanning the first 1000 reporting orders, using the same operational selector. Because the selector is deterministic, the comparison is seed-independent and is run once per generator. The realized greedy outcome with OG-EMS is 0.061 higher than with Base-EMS, corresponding to a 15.1\% relative improvement. This difference is an empirical exposure proxy, not an upper bound on the performance attainable with either generator.}
\label{tab:generator_exposure}
\end{table}

\section{Inference-Time Lookahead: Full Results}
\label{app:lookahead}

\begin{table}[!tb]
\centering
{%
\begin{tabular}{lccc}
\toprule
Group & $N$ orders & Baseline eff. & $\Delta$ items  \\
\midrule
Helped & 226 & 0.72 & $+6.83$ \\
Hurt & 125 & 0.82 & $-5.47$ \\
Neutral & 649 & 0.96 & $0.00$ \\
\bottomrule
\end{tabular}
}
\caption{Lookahead outcome groups over the first 1000 of the 1500 reporting orders. Baseline packing efficiency is the clearest observed separator between hurt and helped groups (0.82 versus 0.72), although this association is not a causal mechanism test.}
\label{tab:lookahead_full}
\end{table}

\paragraph{Full mechanism.}
At each decision step, the method shortlists the top $k=5$ candidates under
the deterministic selector and simulates $d_{\mathrm{LA}}=2$ further
placements from each, using the same selector. Every rollout is scored by a
future-space-quality term
\begin{equation}
\begin{split}
Q={}&
8.0\,f+2.5\,\bar s+1.2\,\bar s_{\rm side}+0.08\,n_{\rm feas}\\
&{}-2.8\,\hat h_{\max}-0.02\,n_{\rm sliv}-0.005\,n_{\rm ems},
\end{split}
\end{equation}
where $f$ is the raw absolute density of the rolled-out layout, $\bar s$ and
$\bar s_{\rm side}$ are its mean bottom and side support, $\hat h_{\max}$ is
its maximum packing height normalized by $H$, $n_{\rm feas}$ is the number
of feasible actions available for the next item, $n_{\rm ems}$ is the number
of raw EMS regions, and $n_{\rm sliv}$ counts those regions satisfying
$0<\mathrm{res}_x<\varepsilon_{\mathrm{sliv}}$ or
$0<\mathrm{res}_y<\varepsilon_{\mathrm{sliv}}$, so exact fits are excluded.
$Q$ is added to the immediate deterministic score with unit weight. These
coefficients were set by hand for this exploratory diagnostic and were not
tuned. The procedure is a heuristic beam-style rollout
\citep{bertsekas2020rollout,owmorton1988filtered}, not a rollout driven by the trained policy or critic.

\paragraph{Outcome groups.}
Table~\ref{tab:lookahead_full} groups orders by the change in absolute density under lookahead relative to the Greedy OG-EMS baseline. \emph{Helped} orders have a positive change, \emph{hurt} orders have a negative change, and \emph{neutral} orders have no change. The mean change in placed-item count is reported as a descriptive characteristic of each group; it does not define group membership.

\paragraph{Why it helps or hurts.}
Orders for which the greedy baseline already has high packing efficiency are more likely to be hurt by lookahead, whereas orders with lower baseline efficiency are more likely to benefit. Across the 1000-order subset, mean absolute density increases from 0.46 without lookahead to 0.48 with lookahead, a gain of 0.014. Item-volume heterogeneity and product-group count do not clearly separate the outcome groups, whereas baseline packing efficiency does. This is an association rather than evidence of a causal threshold mechanism.

\paragraph{Lookahead over the learned policy.}
We also test two exploratory variants that apply lookahead after policy-based shortlisting. The actor-driven variant shortlists candidates by the policy logits, while the critic-preranked variant uses one-step value estimates before applying the same future-space-quality score. Because both variants require additional policy evaluations per decision, we evaluate them without retraining on 50 orders for seed 5, using $k=3$ shortlisted candidates and lookahead depth $d_{\mathrm{LA}}=2$.

As shown in Table~\ref{tab:policy_lookahead}, both exploratory conditions have positive mean differences, and the critic-preranked condition has the larger observed difference. Neither is statistically confirmed at $n_{\mathrm{test}}=50$; a full multi-seed evaluation is required before drawing a comparative conclusion.

\begin{table}[t]
\centering
\setlength{\tabcolsep}{3pt}
\renewcommand{\arraystretch}{1.05}
\begin{tabular}{lccc}
\toprule
Condition & Abs.\ dens. & Helped/hurt/neutral & $p$ \\
\midrule
No lookahead     & 0.51 & \multicolumn{1}{c}{--} & \multicolumn{1}{c}{--} \\
Actor lookahead  & 0.52 & 8/5/37    & 0.42 \\
Critic lookahead & 0.53 & 10/4/36   & 0.14 \\
\bottomrule
\end{tabular}
\caption{Exploratory learned-policy lookahead results over 50 orders for seed 5. Each lookahead condition is compared per order against the no-lookahead baseline using a two-sided Wilcoxon signed-rank test without multiple-comparison correction.}
\label{tab:policy_lookahead}
\end{table}

\section{Pallet-Footprint Sensitivity: Full Methodology}
\label{app:pallet}

\paragraph{Compatibility verification.}
Pallet dimensions are parameterized at evaluation time, and the candidate features of Supplementary Material~\ref{app:candidate_features} are recomputed in the active environment. The heightmap encoder uses adaptive average pooling to produce a fixed-size representation. Alternative footprints therefore require no architectural changes, although the numerical ranges of the unnormalized geometric features change with the pallet dimensions.

\paragraph{Target footprints and feasibility.}
Every reporting order passes the item-level footprint screen at 1200$\times$1000\,mm and 800$\times$600\,mm. At 600$\times$400\,mm, an order is retained only if every item fits in at least one horizontal orientation, leaving 1463 of the 1500 reporting orders (97.5\%) and 8228 of the 8503 training orders (96.8\%) from which the reporting set is held out. This screen establishes individual item fit rather than complete-order packability. The 600$\times$400\,mm row of Table~\ref{tab:pallet_main} is therefore computed over 1463 orders, whereas the remaining rows use all 1500. OPAL and OPAL w/ Base-EMS share the remaining architecture and evaluation settings.

\section{Robustness to Sequence Ordering}
\label{app:robustness}

The primary configuration presents items in descending footprint area (Section~\ref{sec:problemformulation}), matching the CUT-1/CUT-2 convention \citep{zhao2022pct,xiong2024gopt}. We test dependence on this ordering by evaluating OPAL without retraining on the same 1500 orders under two alternatives: a fixed random permutation and the reverse of the primary sequence. Absolute density decreases from 0.491 under the primary ordering to 0.445 under random ordering ($-0.046$, a relative decrease of $9.4\%$) and 0.450 under reversed ordering ($-0.041$, $8.4\%$). OPAL is therefore sensitive to item presentation order.

\end{document}